\documentclass{article}

\usepackage{arxiv}

\usepackage[utf8]{inputenc} 
\usepackage[T1]{fontenc}    
\usepackage{url}            
\usepackage{booktabs}       
\usepackage{amsfonts}       
\usepackage{nicefrac}       
\usepackage{microtype}      
\usepackage{lipsum}		

\usepackage{amsmath}
\usepackage{algorithm}
\usepackage{algorithmic}
\usepackage{booktabs}
\usepackage{graphicx}
\usepackage{subfigure}
\usepackage{multirow}

\usepackage{fancyhdr}
\title{Modeling Winner-Take-All Competition in Sparse Binary Projections}

\date{} 					

\author{
  Wenye Li \\
  School of Science and Engineering \\
  The Chinese University of Hong Kong, Shenzhen\\
  Shenzhen, China \\
  \texttt{wyli@cuhk.edu.cn} \\
}

\begin{document}
\maketitle

\begin{abstract}
Inspired by the advances in biological science, the study of sparse binary projection models has attracted considerable recent research attention. The models project dense input samples into a higher-dimensional space and output sparse binary data representations after the Winner-Take-All competition, subject to the constraint that the projection matrix is also sparse and binary. Following the work along this line, we developed a supervised-WTA model when training samples with both input and output representations are available, from which the {\em optimal} projection matrix can be obtained with a simple, effective yet efficient algorithm. We further extended the model and the algorithm to an unsupervised setting where only the input representation of the samples is available. In a series of empirical evaluation on similarity search tasks, the proposed models reported significantly improved results over the state-of-the-art methods in both search accuracies and running speed. The successful results give us strong confidence that the work provides a highly practical tool to real world applications.
\end{abstract}


\section{Introduction}
\label{sec:introduction}

Random projection has emerged as a powerful tool in data analysis applications \cite{bingham2001random}. It is often used to reduce the dimension of data samples in the Euclidean space. It provides a simple and computationally efficient way to reduce the storage complexity of the data by trading a controlled amount of representation error for faster processing speed and smaller model sizes \cite{johnson1984extensions}.

Very recently, with strong biological evidence, a sparse binary projection model called the FLY algorithm was designed and attracted people's much attention. Instead of performing dimension reduction, the algorithm increases the dimension of the input samples with a random sparse binary projection matrix. After the winner-take-all (WTA) competition that happens in the output space, the samples are converted into a set of sparse binary vectors. In similarity search tasks, it was reported that such sparse binary vectors outperformed the hashed vectors produced by the classical locality sensitive hashing (LSH) method that is based on the random dense projection \cite{dasgupta2017neural}.

Following the work along this line, we proposed two models with the explicit treatment of the WTA competition. Instead of residing on the random generation of the projection matrix, one of our models seeks the {\em optimal} projection matrix under a supervised setting, while the other model operates purely in an unsupervised manner. For each model, we derived an algorithm that is surprisingly simple. In empirical evaluations, both algorithms reported significantly improved results in similarity search accuracies and running speed over the state-of-the-art approaches, and hence provided a practical tool in data analysis applications with high potential.

A note on notation. Unless specified otherwise, a capital letter, such as $W$, denotes a matrix. A lower-cased letter, with or without a subscript, denotes a vector or a scalar. For example $w_{i.}$ denotes the $i$-th row, $w_{.j}$ denotes the $j$-th column, and $w_{ij}$ denotes the $\left(i,j\right)$-th entry of the matrix $W$.

The paper is organized as follows. Section \ref{sec:related} introduces the necessary background. Section \ref{sec:model} presents our models and the algorithms. Section \ref{sec:evaluation} reports the experiments and the results, followed by the conclusion in Section \ref{sec:conclustion}.

\section{Background}
\label{sec:related}

\subsection{Sparse Binary Projection Algorithms}
\label{sec:related:projection}

Different from classical projection methods that commonly map data from a higher-dimensional space to a lower-dimensional space, the FLY algorithm increases the dimension of the data. It was designed by simulating the fruit fly's olfactory circuit, whose function is to associate similar odors with similar tags. Each odor is initially represented as a $50$-dimensional feature vector of firing rates. To associate each odor with a tag involves three steps. Firstly, a divisive normalization step \cite{olsen2010divisive} centers the mean of the feature vector. Secondly, the dimension of the feature vector is expanded from $50$ to $2,000$ with a sparse binary connection matrix \cite{caron2013random,zheng2018complete}, which has the same number of ones in each row. Thirdly, the WTA competition is involved as a result of strong inhibitory feedback coming from an inhibitory neuron. After the competition, all but the highest-firing $5\%$ out of the $2,000$ features are silenced \cite{stevens2015fly}. These remaining $5\%$ features just correspond to the tag assigned to the input odor.

The FLY algorithm can be studied as a special form of the LSH method which produces similar hashes for similar input samples. But different from the classical LSH method which reduces the data dimension, the FLY algorithm increases the dimension with a random sparse binary matrix, while ensuring the sparsity and binarization of the data in the output space. Empirically, the FLY algorithm reported improved results over the LSH method \cite{dasgupta2017neural} in similarity search applications. 

The success of the FLY algorithm inspired considerable research attention, among which one of particular interest to us is the LIFTING algorithm \cite{li2018fast} that removes the randomness assumption of the projection matrix, which is partially supported by most recent biological discoveries \cite{zheng2018complete}. In the work, the projection matrix is obtained through supervised learning. Suppose training samples with both dense input representation $X\in \mathcal{R}^{d\times n}$ and sparse output representation $Y\in \left\{0,1\right\}^{d^{\prime}\times n}$ are available. The LIFTING algorithm seeks the projection matrix $W$ that minimizes $\left\Vert WX-Y\right\Vert_{F}^{2}+\beta\left\Vert W\right\Vert_{\frac{1}{2}}$ in the feasible region of sparse binary matrices. To solve the optimization problem, the Frank-Wolfe algorithm was found to have quite good performance \cite{frank1956algorithm,jaggi2013revisiting}.

\subsection{Winner-Take-All Competition}
\label{sec:related:wta}

Evidences in neuroscience showed that excitation and inhibition are common activities in neurons \cite{stevens2015fly,turner2008olfactory}. Based on the lateral information, some neurons raise to the excitatory state, while the others get inhibited and remain silent. Such excitation and inhibition result in competitions among neurons. Modeling neuron competitions is of key importance, with which useful applications are found in a variety of tasks \cite{arbib2003handbook,maass2000computational}. Specifically in machine learning, the competition mechanism has motivated the design of computer algorithms for a long time, from the early self-organizing map \cite{kohonen1990self} to more recent work in developing novel neural network architectures \cite{panousis2018nonparametric,lynch2019winner}.

To model the competition stage, the WTA model is routinely adopted. We are interested with the following form of the WTA model. For a $d$-dimensional input vector $x$ and a given hash length $k\left(k\ll d\right)$, a function $WTA_{k}^{d}:\mathcal{R}^{d}\rightarrow \left\{0,1\right\}^{d}$ outputs a vector $y=WTA_{k}^{d}\left(x\right)$ satisfying, for each $1\le i\le d$,
\begin{equation}
y_i=\left\{ 
\begin{array}{cl}
1, & \mbox{if $x_i$ is among top-$k$ entries of $\left(x_1,\cdots,x_d\right)$.}\\ 
0, & \mbox{otherwise.}%
\end{array}%
\right. 
\label{equ:wta}
\end{equation}
Thus the output entries with value $1$ just mark the positions of top-$k$ values of $x$. For simplicity and without causing ambiguity, we do not differentiate whether the input/output vector of the $WTA$ function is a row vector or a column vector.

\section{Model}
\label{sec:model}

\subsection{Supervised Training}
\label{sec:model:supervised}

We start from a supervised setting. Let a set of samples be given in the form of $X\in \mathcal{R}^{d\times n}$ and $Y\in \left\{0,1 \right\}^{d^{\prime}\times n}$ with each $x_{.m}\left(1\le m\le n\right)$ being an input sample and $y_{.m}$ being its output representation satisfying $\left\|y_{.m}\right\|_{1}=k$ for a given integer $k$. We assume that, for a fixed integer $c$ \footnote{As in \cite{dasgupta2017neural}, $c$ is set to $\left\lfloor 0.1\times d \right\rfloor$ in this paper.}, there exists a projection matrix $W\in \left\{0,1\right\}^{d^{\prime}\times d}$ with $\left\|w_{i.}\right\|_{1}=c$ ($1\le i\le d^{\prime}$) and $y_{.m}=WTA_{k}^{d^{\prime}}\left(Wx_{.m}\right)$.

The WTA function defined in Eq. (\ref{equ:wta}) satisfies:
\begin{equation}
w_{i.}x_{.m}\ge w_{j.}x_{.m}, \mbox{ if $y_{im}=1$ and $y_{jm}=0$}
\label{equ:ktwa2}
\end{equation} 
for all $1\le m\le n$ and $1\le i,j\le d^{\prime}$.

Now we are interested in inferring such a projection matrix $W$ from the given data. But unfortunately, seeking the matrix directly from Eq. (\ref{equ:ktwa2}) is generally hard. A matrix that satisfies all the constraints may not exist due to the noise in the observed samples. Even if it exists, the computational requirement can be non-trivial. A straightforward modeling of the problem as a linear integer program would involve $d^{\prime}\times d$ variables and $O\left(nk\left(d^{\prime}-k\right)+d^{\prime}\right)$ constraints, which is infeasible to solve even for moderately small $n$ and $d^{\prime}$.

To ensure the tractability, we resort to a relaxation approach. For any feasible $m$, $i$ and $j$, we define a measure $y_{im}\left(1-y_{jm}\right)\left(w_{i.}x_{.m}-w_{j.}x_{.m}\right)$ to quantify the compliance with the condition in Eq. (\ref{equ:ktwa2}). When $y_{im}=1$ and $y_{jm}=0$, the measure is non-negative if the condition is met; otherwise, it is negative. Naturally, we sum up the values of the measure over all $m$, $i$ and $j$, and define
\begin{equation}
L_{s}\left(W\right) = \sum_{m=1}^{n}\sum_{i=1}^{d^{\prime}}\sum_{j=1}^{d^{\prime}}y_{im}\left(1-y_{jm}\right)\left(w_{i.}x_{.m}-w_{j.}x_{.m}\right).
\label{equ:supervised}
\end{equation}

The value of $L_s\left(W\right)$ measures how well a matrix $W$ meets the conditions in Eq. (\ref{equ:ktwa2}). Maximizing $L_s$ with respect to $W$ in the feasible region of sparse binary matrices provides a principled solution to seeking the projection matrix. And we call it the supervised-WTA model.

Considering that
\begin{align*}
& \max L_{s}\left(W\right) \\
\iff &\max \sum_{m=1}^{n}\left[ d^{\prime}\sum_{i=1}^{d^{\prime}}y_{im}w_{i.}x_{.m}-k\sum_{j=1}^{d^{\prime}}w_{j.}x_{.m}\right] \\
\iff &\max \sum_{m=1}^{n}\left[\sum_{i=1}^{d^{\prime}}y_{im}w_{i.}x_{.m}-\frac{k}{d^{\prime}}\sum_{i=1}^{d^{\prime}}w_{i.}x_{.m}\right] \\
\iff &\max \sum_{m=1}^{n}\left[\sum_{i=1}^{d^{\prime}}\left(y_{im}-\frac{k}{d^{\prime}}\right)w_{i.}x_{.m}\right] \\
\iff &\sum_{i=1}^{d^{\prime}}\max \left\{w_{i.}\left[ \sum_{m=1}^{n}x_{.m}\left(y_{im}-\frac{k}{d^{\prime}}\right) \right]\right\} \\
\end{align*}%
Therefore, maximizing $L_{s}\left(W\right)$ is equivalent to $d^{\prime}$ maximization sub-problems. Each sub-problem seeks a row vector $w_{i.}\left(1\le i\le d^{\prime}\right)$ by
\begin{equation}
\max w_{i.}\left[\sum_{m=1}^{n}x_{.m}\left(y_{im}-\frac{k}{d^{\prime}}\right) \right]
\label{equ:subproblem}
\end{equation}
subject to:
\begin{equation}
\mbox{$w_{i.}\in \left\{0,1\right\}^{1\times d}$, and $\left\|w_{i.} \right\|_{1}=c$.}
\label{equ:wconstraint}
\end{equation}

Denote
\begin{equation}
\ell_{.i}=\sum_{m=1}^{n}x_{.m}\left(y_{im}-\frac{k}{d^{\prime}}\right),
\label{equ:solutionl}
\end{equation} 
and the optimal solution of $w_{i.}$ to Eq. (\ref{equ:subproblem}) is given by
\begin{equation}
w^{*}_{i.}=WTA_{c}^{d}\left(\ell_{.i}\right).
\label{equ:solutionw}
\end{equation}

\subsection{Unsupervised Training}
\label{sec:model:unsupervised}

The supervised-WTA model utilizes both input and output representations to learn a projection matrix. When only the input representation is available, we can extend the work to an unsupervised-WTA model, by maximizing the objective:
\begin{equation}
L_{u}\left(W,Y\right) = \sum_{m=1}^{n}\sum_{i=1}^{d^{\prime}}\sum_{j=1}^{d^{\prime}}y_{im}\left(1-y_{jm}\right)\left(w_{i.}x_{.m}-w_{j.}x_{.m}\right)
\label{equ:unsupervised}
\end{equation}
subject to the constraints:  
$w_{i.}\in \left\{0,1\right\}^{1\times d}$, $\left\|w_{i.} \right\|_{1}=c$, $y_{.m}\in \left\{0,1\right\}^{d^{\prime}\times 1}$, and $\left\|y_{.m} \right\|_{1}=k$ 
for all $1\le i\le d^{\prime}$ and $1\le m\le n$.

Different from the supervised model, the unsupervised model treats the unknown output representation $Y$ as a variable, and jointly optimizes on both $W$ and $Y$. To maximize $L_{u}$, an alternating algorithm can be used. Start with a random initialization of $W$ as $W^{1}$, and solve the model iteratively. In $t$-th ($t=1,2,\cdots$) iteration, maximize $L_{u}\left(W^{t},Y\right)$ with respect to $Y$ and get the optimal $Y^{t}$. Then maximize $L_{u}\left(W,Y^{t}\right)$ with respect to $W$ and get the optimal $W^{t+1}$.

The optimal $Y^{t}$ is given by:
\begin{equation}
y^{t}_{.m}=WTA_{k}^{d^{\prime}}\left(W^{t}x_{.m}\right)
\label{equ:solutionyt}
\end{equation}
for all $1\le m\le n$. Similarly to the supervised model, the optimal $W^{t+1}$ is given by:
\begin{equation}
w^{t+1}_{i.}=WTA_{c}^{d}\left(\ell_{.i}^{t}\right)
\label{equ:solutionwt}
\end{equation}
for all $1\le i\le d^{\prime}$, where $\ell_{.i}^{t}=\sum_{m=1}^{n}x_{.m}\left(y_{im}^{t}-\frac{k}{d^{\prime}}\right)$.

Denote by $L_{u}^{t}=L_{u}\left(W^{t},Y^{t}\right)$. Obviously, the sequence $\left\{L_{u}^{t}\right\}$ monotonically increases for $t=1,2,\cdots$. Therefore the alternating optimization process is guaranteed to converge when the objective value $L_{u}^{t}$ can't be increased any more.


It is worth mentioning that the unsupervised-WTA model can be studied as a generic clustering method \cite{jain1999data}. The model puts $m$ data samples into $d^{\prime}$ clusters and each sample belongs to $k$ clusters. A special case of $k=1$ leads to a hard clustering method. Two samples with the element of one in the same output dimension indicates that they have the same cluster membership.

The unsupervised-WTA model can also be treated as a feature selection method \cite{guyon2003introduction}. This can be seen from the fact that each output dimension is associated with a subset of $c$ features, instead of all $d$ features in the input space. The model is able to choose these $c$ features automatically and encode the information in the projection matrix $W$. A detailed discussion of the clustering and the feature selection viewpoints goes beyond the scope of this paper and is hence omitted.

\subsection{Complexity Issues}
\label{sec:model:complexity}

Computing the optimal solution to the supervised-WTA model is straightforward and can be implemented with high efficiency. To obtain each projection vector $w_{i.}$, a na\"ive implementation needs $O\left(dn+d\log c\right)$ operations, among which $O\left(dn\right)$ are for the summation operation in Eq. (\ref{equ:solutionl}) and $O\left(d\log c\right)$ are for the sorting operations in Eq. (\ref{equ:solutionw}) by the Heapsort algorithm \cite{knuth1998art}. Therefore, computing the whole projection matrix needs $O\left(d^{\prime}dn+d^{\prime}d\log c\right)$ operations. In fact, by utilizing the sparse structure of the output matrix $Y$, the computational complexity for $W$ can be further reduced to $O\left(kdn+d^{\prime}d\log c\right)$. As seen in Section \ref{sec:evaluation:speed}, this is a highly efficient result.

To solve the unsupervised-WTA model, in each iteration we need to compute both $Y$ and $W$. Computing one $Y$ needs $O\left(cdn+d^{\prime}d\log k\right)$ operations by utilizing the sparse structure of $W$, where $O\left(cdn\right)$ are for multiplying $W$ with $X$ and $O\left(d^{\prime}d\log k\right)$ are for the sorting operations in Eq. (\ref{equ:solutionyt}). Computing one $W$ has the same complexity as in the supervised-WTA model, $O\left(kdn+d^{\prime}d\log c\right)$. Therefore, the total complexity per iteration is $O\left(\left(k+c\right)dn+d^{\prime}d\log \left(kc\right)\right)$, which is also an efficient solution as seen in Section \ref{sec:evaluation:speed}.

For both WTA models, the memory requirement is mainly from the matrices $X$, $Y$ and $W$, and the storage complexity is $O\left(dn+d^{\prime}n+d^{\prime}d\right)$, which can be further reduced to $O\left(dn+kn+d^{\prime}c\right)$ if sparse matrix representation is adopted.

The training algorithms are parallelizable. Each vector of $W$ and $Y$ can be solved independently with high parallel efficiency. It is also notable that, after simple pre-processing of the training data, all computations only involve simple vector addition and scalar comparison operations.

\section{Evaluation}
\label{sec:evaluation}

\begin{table*}[ht]
  \tiny
  \caption{\textbf{Search accuracies on various datasets with fixed output dimension ($d^{\prime}=2,000$)}. On {\em ImageNet} with one million samples, the results of SUP/LIFTING algorithms are not available due to the prohibitive computation to obtain the output representations.}
  \label{tab:comparison}
  \vspace{1.5mm}
  \centering
  \begin{tabular}{l|c|c|c|c|c|c|c|c|c|c}
    \hline \hline
    Datasets & $k$ & {\em SUP} & {\em UNSUP} & {\em LSH} & {\em FJL} & {\em FLY} & {\em LIFTING} & {\em ITQ} & {\em SPH} & {\em ISOH} \\
    \hline
    & 2 & $\mathbf{0.1758}$ & $0.1143$ & $0.0174$ & $0.0169$ & $0.0474$ & $0.1748$ & $0.0103$ & $0.0097$ & $0.0101$ \\ \cline{2-11}
    {\em ARTFC} & 4 & $\mathbf{0.6665}$ & $0.3531$ & $0.0243$ & $0.0237$ & $0.0673$ & $0.6134$ & $0.0175$ & $0.0138$ & $0.0227$ \\ \cline{2-11}
    $d=1,000$ & 8 & $0.3647$ & $\mathbf{0.3944}$ & $0.0259$ & $0.0255$ & $0.0376$ & $0.2612$ & $0.0360$ & $0.0173$ & $0.0331$ \\ \cline{2-11}
    & 16 & $\mathbf{0.5884}$ & $0.3267$ & $0.0278$ & $0.0282$ & $0.0402$ & $0.1694$ & $0.0367$ & $0.0202$ & $0.0349$ \\ \cline{2-11}
    & 32 & $\mathbf{0.3141}$ & $0.1319$ & $0.0324$ & $0.0336$ & $0.0443$ & $0.0832$ & $0.0382$ & $0.0235$ & $0.0375$ \\ 
    \hline\hline
    & 2 & $0.1317$ & $\mathbf{0.1596}$ & $0.0217$ & $0.0198$ & $0.0511$ & $0.0831$ & $0.0221$ & $0.0195$ & $0.0198$ \\ \cline{2-11}
    {\em GLOVE} & 4 & $0.2310$ & $\mathbf{0.3251}$ & $0.0356$ & $0.0328$ & $0.0964$ & $0.1458$ & $0.0617$ & $0.0311$ & $0.0594$ \\ \cline{2-11}
    $d=300$ & 8 & $0.3061$ & $\mathbf{0.3959}$ & $0.0655$ & $0.0618$ & $0.1073$ & $0.1914$ & $0.1209$ & $0.0591$ & $0.1112$ \\ \cline{2-11}
    & 16 & $0.4030$ & $\mathbf{0.4495}$ & $0.1138$ & $0.1081$ & $0.1809$ & $0.2851$ & $0.1939$ & $0.1004$ & $0.1882$ \\ \cline{2-11}
    & 32 & $\mathbf{0.4374}$ & $0.4323$ & $0.2039$ & $0.2139$ & $0.2808$ & $0.3917$ & $0.3208$ & $0.1717$ & $0.2727$ \\ 
    \hline\hline
    & 2 & $\mathbf{0.2860}$ & $0.2476$ & $0.0369$ & $0.0422$ & $0.1119$ & $0.1159$ & $0.0288$ & $0.0237$ & $0.0194$ \\ \cline{2-11}
    {\em MNIST} & 4 & $0.3338$ & $\mathbf{0.3829}$ & $0.0844$ & $0.1029$ & $0.1721$ & $0.2003$ & $0.0852$ & $0.0649$ & $0.0773$ \\ \cline{2-11}
    $d=784$ & 8 & $0.3885$ & $\mathbf{0.4387}$ & $0.1823$ & $0.2004$ & $0.2717$ & $0.3044$ & $0.2008$ & $0.1443$ & $0.1601$ \\ \cline{2-11}
    & 16 & $0.4698$ & $\mathbf{0.4957}$ & $0.3226$ & $0.3409$ & $0.3953$ & $0.4150$ & $0.3207$ & $0.2559$ & $0.3101$ \\ \cline{2-11}
    & 32 & $0.5108$ & $\mathbf{0.5207}$ & $0.4773$ & $0.4846$ & $0.5162$ & $0.5130$ & $0.4415$ & $0.3616$ & $0.4067$ \\ 
    \hline\hline
    & 2 & $\mathbf{0.1706}$ & $0.1502$ & $0.0355$ & $0.0349$ & $0.1066$ & $0.1139$ & $0.0274$ & $0.0272$ & $0.0230$ \\ \cline{2-11}
    {\em SIFT} & 4 & $0.2240$ & $\mathbf{0.2278}$ & $0.0760$ & $0.0707$ & $0.1592$ & $0.2120$ & $0.0550$ & $0.0596$ & $0.0623$ \\ \cline{2-11}
    $d=128$ & 8 & $0.3768$ & $\mathbf{0.3912}$ & $0.1556$ & $0.1692$ & $0.2382$ & $0.3059$ & $0.0951$ & $0.1153$ & $0.1240$ \\ \cline{2-11}
    & 16 & $0.4353$ & $\mathbf{0.4461}$ & $0.2751$ & $0.2698$ & $0.3409$ & $0.3529$ & $0.1712$ & $0.1993$ & $0.1905$ \\ \cline{2-11}
    & 32 & $\mathbf{0.4839}$ & $0.4751$ & $0.4122$ & $0.4290$ & $0.4504$ & $0.4295$ & $0.3217$ & $0.2582$ & $0.2832$ \\ \hline\hline
    & 2 & {\em N.A.} & $\mathbf{0.1280}$ & $0.0251$ & $0.0224$ & $0.0668$ & {\em N.A.} & $0.0197$ & $0.0174$ & $0.0202$ \\ \cline{2-11}
    {\em ImageNet} & 4 & {\em N.A.} & $\mathbf{0.1863}$ & $0.0502$ & $0.0578$ & $0.1058$ & {\em N.A.} & $0.0406$ & $0.0389$ & $0.0392$ \\ \cline{2-11}
    $d=1,000$ & 8 & {\em N.A.} & $\mathbf{0.2177}$ & $0.0824$ & $0.0854$ & $0.1519$ & {\em N.A.} & $0.0925$ & $0.0806$ & $0.0826$ \\ \cline{2-11}
    & 16 & {\em N.A.} & $\mathbf{0.2391}$ & $0.1522$ & $0.1527$ & $0.2122$ & {\em N.A.} & $0.1679$ & $0.1338$ & $0.1378$ \\ \cline{2-11}
    & 32 & {\em N.A.} & $\mathbf{0.2480}$ & $0.2282$ & $0.2337$ & $0.2430$ & {\em N.A.} & $0.2311$ & $0.1801$ & $0.2002$ \\ \hline \hline
  \end{tabular}
\end{table*}

\begin{table*}[ht]
  \tiny
  \caption{\textbf{Search accuracies on GLOVE dataset with various input dimensions and fixed output dimension ($d^{\prime}=2,000$)}.}
  \label{tab:glove}
  \vspace{1.5mm}
  \centering
  \begin{tabular}{l|c|c|c|c|c|c|c|c|c|c}
    \hline \hline
    Dimension & $k$ & {\em SUP} & {\em UNSUP} & {\em LSH} & {\em FJL} & {\em FLY} & {\em LIFTING} & {\em ITQ} & {\em SPH} & {\em ISOH} \\
    \hline
    \multirow{5}{*}{\em $d=100$}  
    & 2 & $0.1007$ & $\mathbf{0.1210}$ & $0.0208$ & $0.0210$ & $0.0503$ & $0.0982$ & $0.0245$ & $0.0229$ & $0.0230$ \\ \cline{2-11}
    & 4 & $0.1720$ & $\mathbf{0.2274}$ & $0.0335$ & $0.0317$ & $0.0787$ & $0.1449$ & $0.0683$ & $0.0452$ & $0.0633$ \\ \cline{2-11}
    & 8 & $0.2365$ & $\mathbf{0.2816}$ & $0.0591$ & $0.0602$ & $0.1059$ & $0.1898$ & $0.1509$ & $0.0705$ & $0.1297$ \\ \cline{2-11}
    & 16 & $0.3113$ & $\mathbf{0.3572}$ & $0.1096$ & $0.1125$ & $0.1698$ & $0.2279$ & $0.2201$ & $0.1232$ & $0.1995$ \\ \cline{2-11}
    & 32 & $0.3779$ & $\mathbf{0.3831}$ & $0.1962$ & $0.2007$ & $0.2581$ & $0.2954$ & $0.3311$ & $0.2398$ & $0.2952$ \\ \hline\hline
    \multirow{5}{*}{\em $d=200$}  
    & 2 & $0.0808$ & $\mathbf{0.1073}$ & $0.0183$ & $0.0177$ & $0.0387$ & $0.0733$ & $0.0231$ & $0.0223$ & $0.1234$ \\ \cline{2-11}
    & 4 & $0.1432$ & $\mathbf{0.2030}$ & $0.0275$ & $0.0257$ & $0.0624$ & $0.1037$ & $0.0692$ & $0.0395$ & $0.0212$ \\ \cline{2-11}
    & 8 & $0.2008$ & $\mathbf{0.2551}$ & $0.0459$ & $0.0329$ & $0.0786$ & $0.1363$ & $0.1197$ & $0.0624$ & $0.1256$ \\ \cline{2-11}
    & 16 & $0.2759$ & $\mathbf{0.3189}$ & $0.0816$ & $0.0798$ & $0.1284$ & $0.1712$ & $0.1804$ & $0.1105$ & $0.1905$ \\ \cline{2-11}
    & 32 & $0.3254$ & $\mathbf{0.3331}$ & $0.1442$ & $0.1502$ & $0.1991$ & $0.2391$ & $0.3025$ & $0.2051$ & $0.2782$ \\ \hline\hline
    \multirow{5}{*}{\em $d=500$}  
    & 2 & $0.0689$ & $\mathbf{0.0866}$ & $0.0148$ & $0.0152$ & $0.0226$ & $0.0490$ & $0.0197$ & $0.0173$ & $0.0182$ \\ \cline{2-11}
    & 4 & $0.1328$ & $\mathbf{0.1702}$ & $0.0195$ & $0.0183$ & $0.0394$ & $0.0711$ & $0.0522$ & $0.0301$ & $0.0397$ \\ \cline{2-11}
    & 8 & $0.1878$ & $\mathbf{0.2252}$ & $0.0278$ & $0.0276$ & $0.0421$ & $0.0892$ & $0.0973$ & $0.0521$ & $0.0885$ \\ \cline{2-11}
    & 16 & $\mathbf{0.2768}$ & $0.2696$ & $0.0437$ & $0.0469$ & $0.0710$ & $0.1188$ & $0.1497$ & $0.0995$ & $0.1305$ \\ \cline{2-11}
    & 32 & $\mathbf{0.3172}$ & $0.2727$ & $0.0728$ & $0.0804$ & $0.1115$ & $0.1806$ & $0.2119$ & $0.1502$ & $0.2117$ \\ \hline\hline
    \multirow{5}{*}{\em $d=1,000$}  
    & 2 & $0.0464$ & $\mathbf{0.0508}$ & $0.0132$ & $0.0129$ & $0.0172$ & $0.0293$ & $0.0177$ & $0.0166$ & $0.0179$ \\ \cline{2-11}
    & 4 & $0.0985$ & $\mathbf{0.1131}$ & $0.0162$ & $0.0175$ & $0.0288$ & $0.0396$ & $0.0356$ & $0.0289$ & $0.0322$ \\ \cline{2-11}
    & 8 & $0.1447$ & $\mathbf{0.1615}$ & $0.0214$ & $0.0261$ & $0.0303$ & $0.0490$ & $0.0434$ & $0.0312$ & $0.0365$ \\ \cline{2-11}
    & 16 & $\mathbf{0.2115}$ & $0.2071$ & $0.0305$ & $0.0372$ & $0.0474$ & $0.0662$ & $0.0912$ & $0.0787$ & $0.0883$ \\ \cline{2-11}
    & 32 & $\mathbf{0.2194}$ & $0.2049$ & $0.0464$ & $0.0511$ & $0.0699$ & $0.0874$ & $0.1507$ & $0.1339$ & $0.1303$ \\ \hline \hline
  \end{tabular}
\end{table*}

\begin{figure*}[ht]
\centering
 \subfigure[ARTFC (1K)]{
   \includegraphics[width=1.2in,height=1.2in]{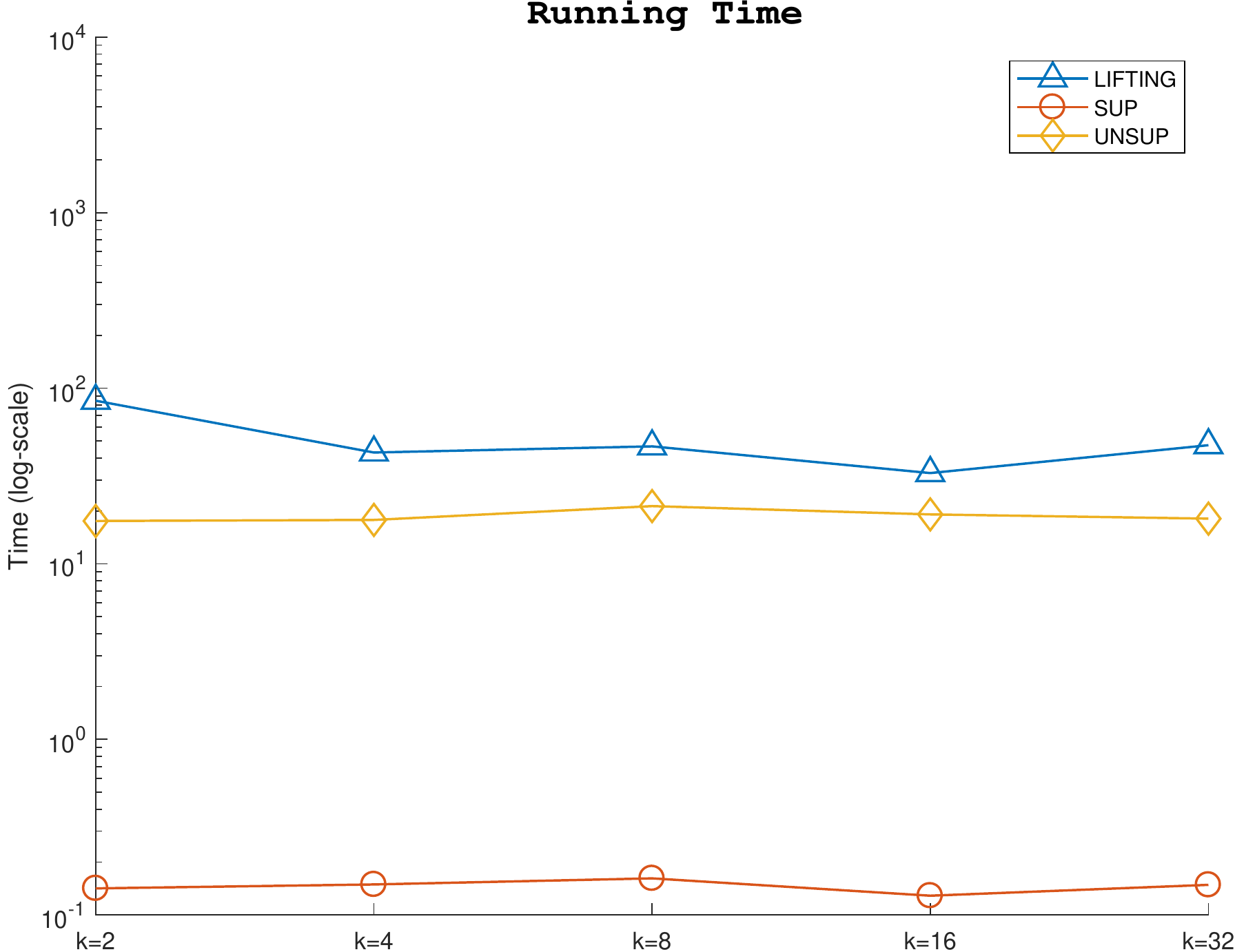}
   \label{fig:speed:art1k}
 }
 \subfigure[ARTFC (5K)]{
   \includegraphics[width=1.2in,height=1.2in]{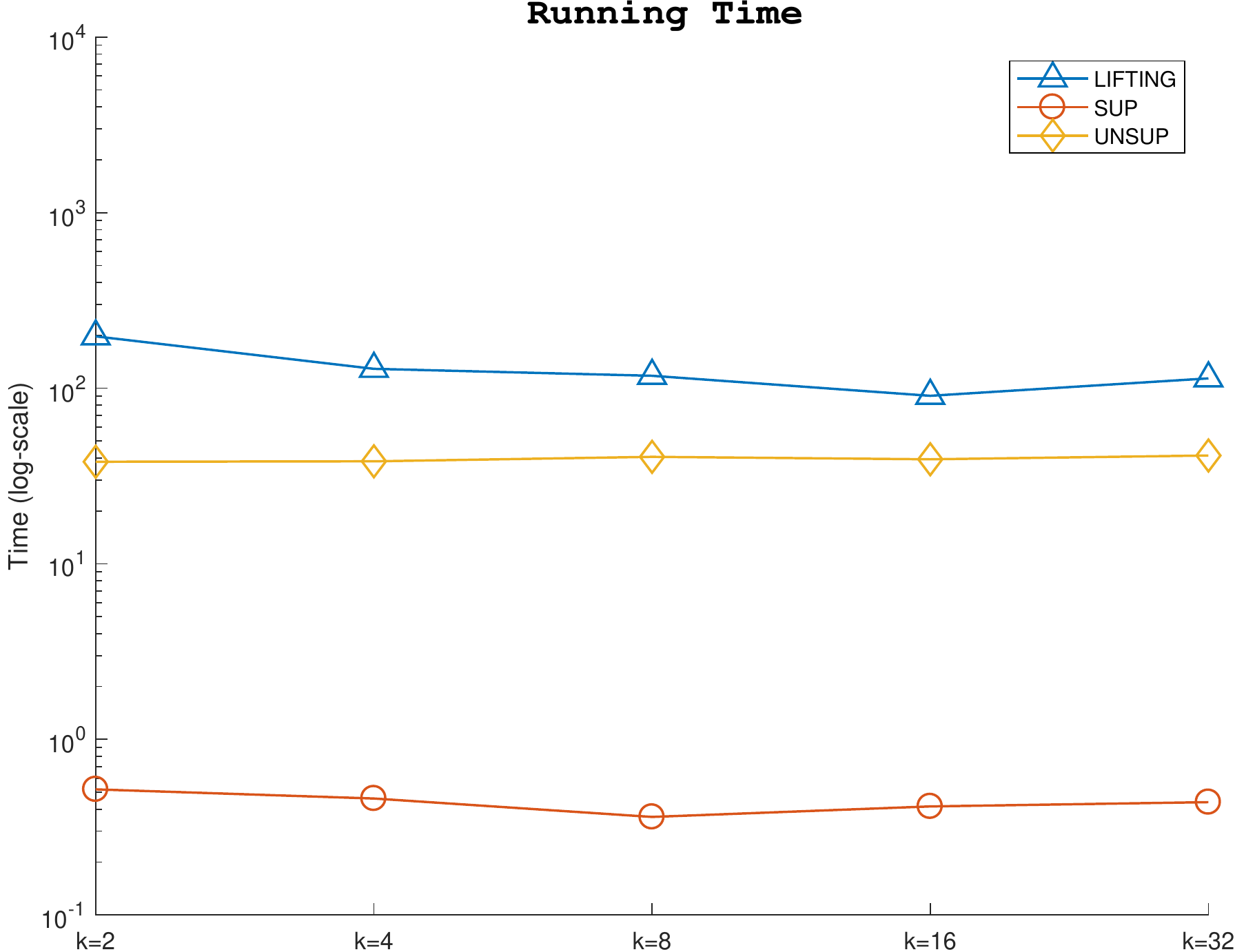}
   \label{fig:speed:art5k}
 }
 \subfigure[ARTFC (10K)]{
   \includegraphics[width=1.2in,height=1.2in]{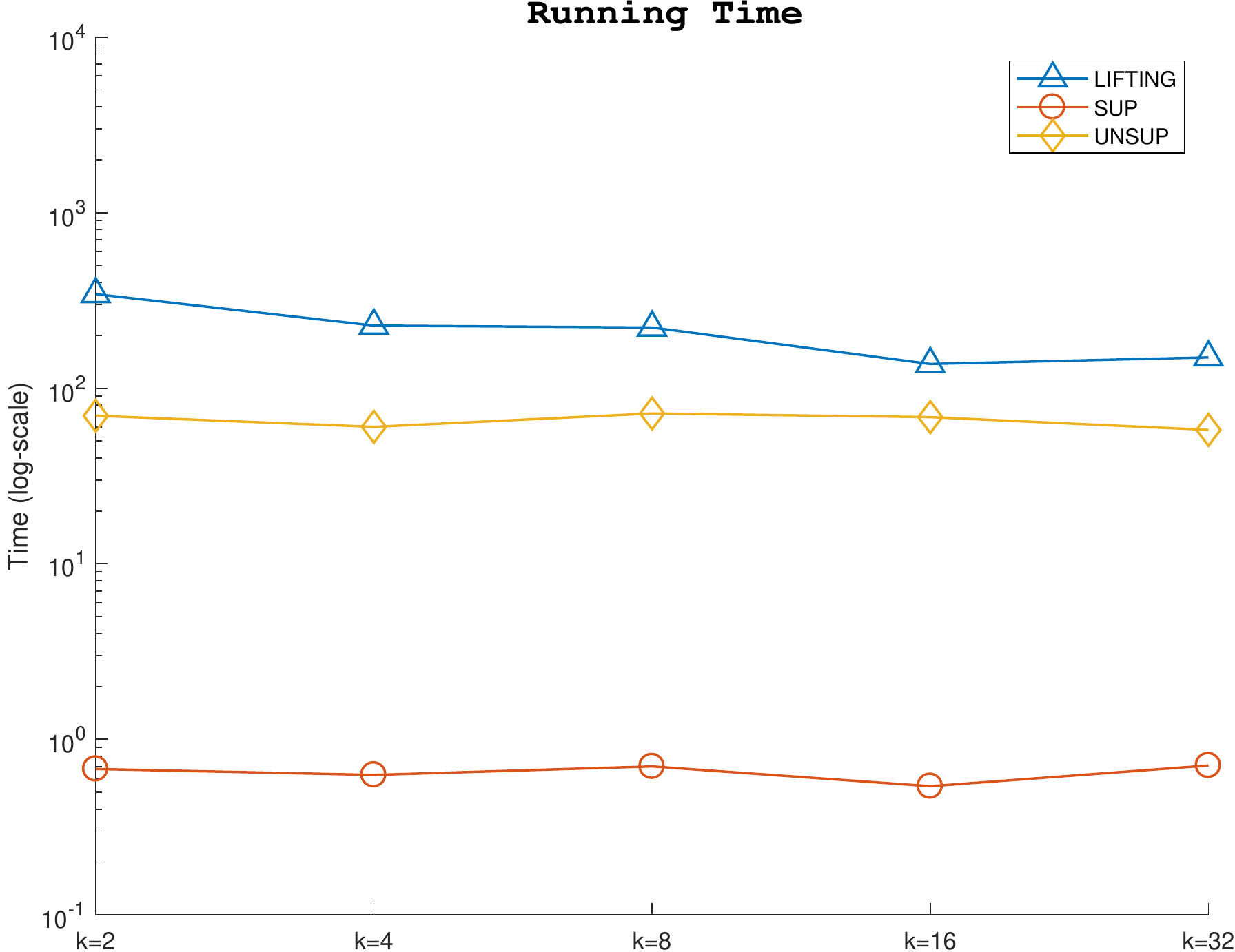}
   \label{fig:speed:art10k}
 } 
 \subfigure[ARTFC (50K)]{
   \includegraphics[width=1.2in,height=1.2in]{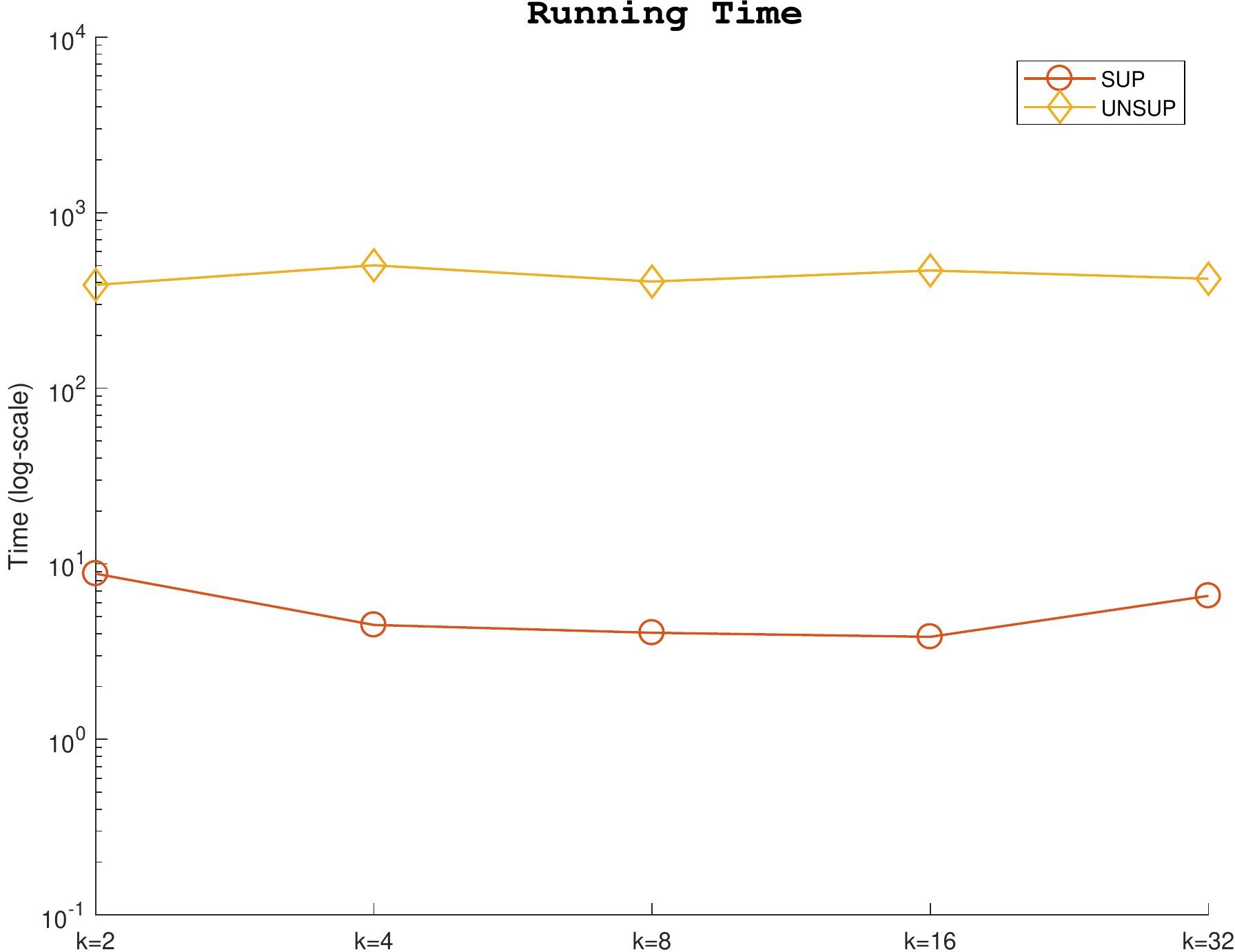}
   \label{fig:speed:art50k}
 }
 \subfigure[ImageNet]{
   \includegraphics[width=1.2in,height=1.2in]{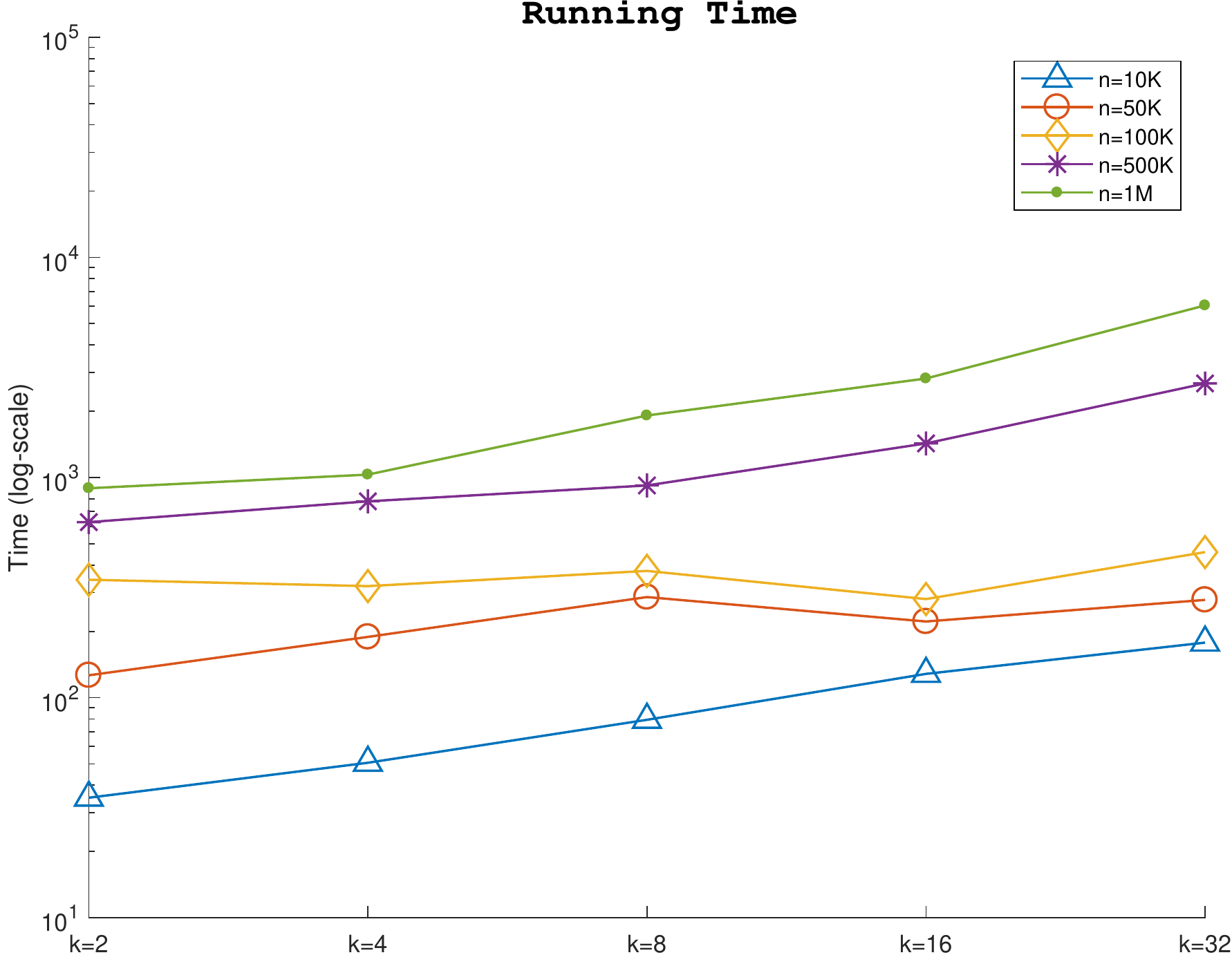}
   \label{fig:speed:imagenet}
 }

\caption{\textbf{Comparison of training time}. Horizontal: the hash length ($k=2/4/8/16/32$). Vertical: training time (seconds) in log-scale. (a)-(d): Comparison of LIFTING/SUP/UNSUP algorithms with $1K$ to $50K$ training samples of the ARTFC dataset (with both input and output representations). In (d), the results of LIFTING are not available due to the prohibitive computation. (e) Training time of the UNSUP algorithm with $10K$ to $1M$ training samples of the ImageNet dataset. The results of SUP/LIFTING algorithms are not available due to the prohibitive computation to obtain the output representations.}
\label{fig:exp:speed}
\end{figure*}

\subsection{General Settings}
\label{sec:evaluation:settings}

To evaluate the performance of the proposed models, we carried out a series of experiments under the following settings.

\textbf{Application:} Similarly to the work of \cite{dasgupta2017neural}, we applied the proposed models in similarity search tasks. Similarity search aims to find similar samples to a given query object among potential candidates, according to a certain distance or similarity measure \cite{baeza1999modern}. The complexity of accurately determining similar samples relies heavily on both the number of candidates and their dimension. Computing the distances seems straightforward, but unfortunately could often become prohibitive if the number of candidates is too large or the dimension of the data is too high.

To handle the difficulty brought by the high dimension of the input data, we can either reduce the data dimension while approximately preserving their pairwise distances, or increase the dimension but confining the data in the output space to be sparse and binary, in the hope of significantly improved search speed with the new representation.

\textbf{Objective:} Our major objective is to evaluate and compare the similarity search accuracies for different algorithms. Each sample in a given dataset was used, in turn, as the query object, and the other samples in the same dataset were used as the search candidates. For each query object, we compared its $100$ nearest neighbors in the output space with its $100$ nearest neighbors in the input space, and recorded the ratio of common neighbors in both spaces. The ratio is averaged over all query objects as the search accuracy of each algorithm. Obviously, a higher similarity search accuracy indicates a better preserving of locality structures from the input space to the output space by the algorithm.

\textbf{Datasets:} In the evaluation, four real datasets and five artificially generated datasets were used. The real datasets have the input representation $X$ only; while the artificial datasets have both the input representation $X$ and the output representation $Y$. Specifically these datasets are:
\begin{itemize}
  \item GLOVE \cite{pennington2014glove}: $100$- to $1000$-dimensional {\em GloVe} word vectors trained on a subset of 330 million tokens from wikimedia database dumps\footnote{https://dumps.wikimedia.org/} with the $50,000$ most frequent words.
  \item ImageNet \cite{russakovsky2015imagenet}: a large collection of images represented as $1,000$-dimensional visual words quantized from SIFT features.
  \item MNIST \cite{lecun1998gradient}: $784$-dimensional images of handwritten digits in gray-scale.
  \item SIFT \cite{jegou2011product}: $128$-dimensional SIFT descriptors of images used for similarity search.
  \item ARTFC: five sets of $1,000$-dimensional dense vectors ($X$) and $2,000$-dimensional sparse binary vectors ($Y$). For each hash length of $k=2/4/8/16/32$, a set of $2,000$-dimensional sparse binary vectors were randomly generated with the hash length. Then the vectors were projected to $1,000$-dimensional dense vectors through principal component analysis. In this way, the samples' pairwise distances are roughly preserved between the input space and the output space; i.e., $\left\|x_{.m}-x_{.m^{\prime}}\right\|^2\approx \left\|y_{.m}-y_{.m^{\prime}}\right\|^2$ for all pairs of samples in the same set. 
\end{itemize}

\textbf{Algorithms to compare:} We compared the proposed supervised-WTA (denoted by SUP) model and the unsupervised-WTA (UNSUP) model with the LSH algorithm \cite{gionis1999similarity,charikar2002similarity}, the fast Jonson-Lindenstrauss projection (FJL) algorithm \cite{ailon2009fast}, the FLY algorithm \cite{dasgupta2017neural} and the LIFTING algorithm \cite{li2018fast}. The LSH algorithm maps $d$-dimensional inputs to $k$-dimensional dense vectors with a random dense projection matrix. The FJL algorithm is a fast implementation of the LSH algorithm with a sparse projection matrix. The FLY algorithm uses a random sparse binary matrix to map $d$-dimensional inputs to $d^{\prime}$-dimensional vectors. The LIFTING algorithm trains a sparse binary projection matrix in a supervised manner for the $d$-dimensional to $d^{\prime}$-dimensional projection. Both the FLY and the LIFTING algorithms involve a WTA competition stage in the output space to generate sparse binary vectors for each hash length.

Besides, we conducted the comparison with a number of other hashing algorithms, including iterative quantization (ITQ) \cite{gong2012iterative}, spherical hashing (SPH) \cite{heo2015spherical} and isotrophic hashing (ISOH) \cite{kong2012isotropic}. These algorithms were popularly used in literature to produce sparse binary data embeddings.

\textbf{Computing environment:} All the algorithms were implemented in MATLAB platform running on an 8-way computing server, with which a maximum of $128$ threads were enabled for each algorithm. For the LIFTING algorithm, IBM CPLEX was used as the linear program solver that was needed by the Frank-Wolfe algorithm.

\subsection{Similarity Search Accuracy}
\label{sec:evaluation:supervised}

We carried out the experiment on the artificial datasets and the real datasets. From each ARTFC dataset, we randomly chose $10,000$ training samples with both the input ($X$) and the output ($Y$) representations, and chose another $10,000$ testing samples with the input representation only. For the two proposed WTA models, we trained a sparse binary projection matrix $W$ each based on the training data. Then we generated $2,000$-dimensional sparse binary output vectors via the WTA competition after projecting the testing samples with the matrix. For the LIFTING algorithm, the same training and testing procedures were applied. For all other algorithms, we applied each of them on the testing samples to get either dense or sparse binary output vectors. Then the output vectors are used in similarity search and compared against the input vectors, as illustrated in Section \ref{sec:evaluation:settings}.

We repeated the process for fifty runs and recorded the average accuracies. The results are given in Table \ref{tab:comparison}. Each row shows the similarity search accuracies with a specific hash length \footnote{As in \cite{dasgupta2017neural,li2018fast}, the hash length is defined as the number of ones in each output vector for the FLY, LIFTING and WTA algorithms. For other algorithms, it is defined as the output dimension.}. Consistent with the results reported in \cite{dasgupta2017neural}, the sparse binary projection algorithms reported improved results over the classical LSH method. Among the algorithms, it is evidently shown that, with the support of the supervised information, the LIFTING and the supervised-WTA algorithms reported further improved results over the FLY algorithm. Most prominently, with the hash length $k=4$, the FLY algorithm has an accuracy of $6.73\%$, while the supervised-WTA model's accuracy reaches $66.7\%$, almost ten times higher. When comparing the two supervised algorithms, the supervised-WTA model outperformed the LIFTING algorithm with all hash lengths.

Among the unsupervised learning algorithms, the proposed unsupervised-WTA model reported the best performances, significantly better than the results given by the LSH, FJL, FLY, ITQ, SPH and ISOH algorithms. Its accuracies are even better than the supervised-WTA model with the hash length $k=8$.

On GLOVE/MNIST/SIFT datasets, only the input representation $X$ is available. We randomly chose $10,000$ samples for training and $10,000$ samples for testing. As suggested in \cite{li2018fast}, we computed $Y^{\ast}=\arg_{Y}\min\frac{1}{2}\left\Vert X^{T}X-Y^{T}Y\right\Vert_{F}^{2}+\gamma\left\Vert Y\right\Vert_{\frac{1}{2}}$ for the training data via the Frank-Wolfe algorithm, and used $Y^{*}$ as the output representation for training. Then we carried out the experiment under the same setting as on ARTFC datasets. Again the two WTA models reported evidently improved results. 

When comparing the two WTA models on these datasets, the unsupervised-WTA model performed even better than the supervised-WTA model on most tests. In cases with known $X$ only, an approximation of $Y$ has to be obtained through matrix factorization. The quality of this approximated $Y$ becomes critical to the supervised-WTA model. We believe this is the major reason why the supervised model no longer excels.

Besides, we tested the algorithms' performances on a much larger ImageNet dataset with one million images for training and $10,000$ images for testing. Computing the output representation $Y^{\ast}$ becomes infeasible on such a large training set, and therefore the results of SUP and LIFTING were not available. Comparing the with the other available algorithms, once again the unsupervised-WTA algorithm reported significantly improved search accuracies.

In addition to the experiment on similarity search accuracies, we further investigated the influence of different input/output dimensions on the performance of the proposed models. We fixed the output dimension to $d^{\prime}=2,000$ while varying the input dimension from $100$ to $1,000$ on GloVe word vectors. We recorded the similarity search accuracies by all the algorithms. From the results in Table \ref{tab:glove}, we can see that the WTA models reported consistently improved results.

\subsection{Running Speed}
\label{sec:evaluation:speed}

As a practical concern, we compared the training time of the proposed WTA models with the LIFTING algorithm. In the experiment, we used the ARTFC datasets with $1,000$-dimensional inputs and $2,000$-dimensional outputs, and the number of training samples varied from $1,000$ to $50,000$. 

We recorded the training time of each algorithm to compute the sparse binary projection matrix $W$. On all training sets, the proposed models reported significantly faster speed than the LIFTING algorithm. With $1,000$ samples (ref. Fig. \ref{fig:speed:art1k}), the supervised-WTA model took less than $0.2$ seconds to get the optimal solution, hundreds of times faster than the LIFTING algorithm which took around $50$ seconds. 

The unsupervised-WTA model needs to solve multiple $W^{t}$ and $Y^{t}$ iteratively. It took $10$ to $20$ seconds with $1,000$ samples, which was slower than the supervised-WTA model but several times faster than the LIFTING algorithm. With $50,000$ samples (ref. Fig. \ref{fig:speed:art50k}), the supervised-WTA model took less than $10$ seconds, and the unsupervised-WTA model took about $400$ seconds to get the solutions. For the LIFTING algorithm, we didn't finish the execution in our platform within $12$ hours. All these real results were consistent with the complexity analysis given in Section \ref{sec:model:complexity}, and justified the running efficiency of the proposed WTA models.

We further carried out the experiment on the much larger ImageNet dataset and reported the results in Fig. \ref{fig:speed:imagenet}. Due to the prohibitive computation to obtain the output representations, only the results of the UNSUP algorithm are available. From the results we can see, with a hash length $k=32$, the algorithm took less than $200$ seconds to train a projection matrix with $10K$ samples, and took around $6,000$ seconds to train with one million samples. We believe that these are reasonably efficient and promising results to practical application scenarios.

\section{Conclusion}
\label{sec:conclustion}

With strong evidence from biological science, the study of sparse binary projection models has attracted much research attention recently. By mapping lower-dimensional dense data to higher-dimensional sparse binary vectors, the models have reported excellent empirical results and proved to be useful in practical applications. 

Sparse binary projections are tightly coupled with WTA competitions. The competition is an important stage for pattern recognition activities that happen in the brain. Accordingly, our work started from the explicit treatment of the competition, and proposed two models to seek the desired projection matrix. Specifically, one model utilizes both input and output representations of the samples, and trains the projection matrix as a supervised learning problem. The other model utilizes the input representation only and trains the matrix in an unsupervised manner, which equips the model with wider application scenarios. For each model, we developed a simple, effective and efficient algorithm. In the evaluation, the models significantly outperformed the state-of-the-art methods, in both search accuracies and running speed.

Our work potentially triggers a number of topics to study. Firstly, the algorithms for both models only involve simple vector addition and scalar comparison operations, which are highly parallelizable. Such characteristics make the computing procedures suitable to be implemented with customized hardware \cite{omondi2006fpga}, which provides a high-throughput and economical solution for large-scale data analysis applications.

Secondly, the unsupervised-WTA model provides a unified framework that combines the clustering and the feature selection techniques. We firmly believe that new applications along this line are possible. Besides, this viewpoint may provide a potential bridge that helps to make clear why the WTA competition leads to algorithms that preserve the locality structures of the data well, as reported in this paper.

Thirdly, there is potential to design new artificial neural network architectures. The WTA competition and the relaxation techniques adopted in this paper can be used as an activation function of the neurons in an artificial neural network. We warmly anticipate future work along this direction \cite{pehlevan2018similarity,lynch2019winner}.

\section{Acknowledgments}

This work was supported by Shenzhen Fundamental Research Fund (JCYJ20170306141038939, KQJSCX20170728162302784).


\begin{thebibliography}{10}

\bibitem{bingham2001random}
E.~Bingham and H.~Mannila.
\newblock Random projection in dimensionality reduction: applications to image
  and text data.
\newblock In {\em SIGKDD}, pages 245--250. ACM, 2001.

\bibitem{johnson1984extensions}
W.~Johnson and J.~Lindenstrauss.
\newblock Extensions of {L}ipschitz mappings into a {H}ilbert space.
\newblock {\em Contemporary Mathematics}, 26(189-206):1, 1984.

\bibitem{dasgupta2017neural}
S.~Dasgupta, C.~Stevens, and S.~Navlakha.
\newblock A neural algorithm for a fundamental computing problem.
\newblock {\em Science}, 358(6364):793--796, 2017.

\bibitem{olsen2010divisive}
S.~Olsen, V.~Bhandawat, and R.~Wilson.
\newblock Divisive normalization in olfactory population codes.
\newblock {\em Neuron}, 66(2):287--299, 2010.

\bibitem{caron2013random}
S.~Caron, V.~Ruta, L.~Abbott, and R.~Axel.
\newblock Random convergence of olfactory inputs in the drosophila mushroom
  body.
\newblock {\em Nature}, 497(7447):113, 2013.

\bibitem{zheng2018complete}
Z.~Zheng, S.~Lauritzen, E.~Perlman, C.~Robinson, et~al.
\newblock A complete electron microscopy volume of the brain of adult
  drosophila melanogaster.
\newblock {\em Cell}, 174(3):730--743, 2018.

\bibitem{stevens2015fly}
C.~Stevens.
\newblock What the fly’s nose tells the fly’s brain.
\newblock {\em Proceedings of the National Academy of Sciences},
  112(30):9460--9465, 2015.

\bibitem{li2018fast}
W.~Li, J.~Mao, Y.~Zhang, and S.~Cui.
\newblock Fast similarity search via optimal sparse lifting.
\newblock In {\em NIPS}, pages 176--184, 2018.

\bibitem{frank1956algorithm}
M.~Frank and P.~Wolfe.
\newblock An algorithm for quadratic programming.
\newblock {\em Naval Research Logistics}, 3(1-2):95--110, 1956.

\bibitem{jaggi2013revisiting}
M.~Jaggi.
\newblock Revisiting {Frank-Wolfe}: Projection-free sparse convex optimization.
\newblock In {\em ICML}, pages 427--435, 2013.

\bibitem{turner2008olfactory}
G.~Turner, M.~Bazhenov, and G.~Laurent.
\newblock Olfactory representations by drosophila mushroom body neurons.
\newblock {\em Journal of Neurophysiology}, 99(2):734--746, 2008.

\bibitem{arbib2003handbook}
M.~Arbib.
\newblock {\em The Handbook of Brain Theory and Neural Networks}.
\newblock MIT Press, 2003.

\bibitem{maass2000computational}
W.~Maass.
\newblock On the computational power of winner-take-all.
\newblock {\em Neural computation}, 12(11):2519--2535, 2000.

\bibitem{kohonen1990self}
T.~Kohonen.
\newblock The self-organizing map.
\newblock {\em Proceedings of the IEEE}, 78(9):1464--1480, 1990.

\bibitem{panousis2018nonparametric}
K.~Panousis, S.~Chatzis, and S.~Theodoridis.
\newblock Nonparametric bayesian deep networks with local competition.
\newblock In {\em ICML}, 2019.

\bibitem{lynch2019winner}
N.~Lynch, C.~Musco, and M.~Parter.
\newblock Winner-take-all computation in spiking neural networks.
\newblock {\em arXiv preprint arXiv:1904.12591}, 2019.

\bibitem{jain1999data}
A.~Jain, N.~Murty, and P.~Flynn.
\newblock Data clustering: A review.
\newblock {\em ACM Computing Surveys}, 31(3):264--323, 1999.

\bibitem{guyon2003introduction}
I.~Guyon and A.~Elisseeff.
\newblock An introduction to variable and feature selection.
\newblock {\em Journal of Machine Learning Research}, 3(Mar):1157--1182, 2003.

\bibitem{knuth1998art}
D.~Knuth.
\newblock {\em The Art of Computer Programming, Sorting and Searching},
  volume~3.
\newblock Addison-Wesley, Reading, 1998.

\bibitem{baeza1999modern}
R.~Baeza-Yates and B.~Ribeiro-Neto.
\newblock {\em Modern Information Retrieval}, volume 463.
\newblock ACM Press, 1999.

\bibitem{pennington2014glove}
J.~Pennington, R.~Socher, and C.~Manning.
\newblock Glove: Global vectors for word representation.
\newblock In {\em EMNLP}, pages 1532--1543, 2014.

\bibitem{russakovsky2015imagenet}
O.~Russakovsky, J.~Deng, H.~Su, et~al.
\newblock Imagenet large scale visual recognition challenge.
\newblock {\em International Journal of Computer Vision}, 115(3):211--252,
  2015.

\bibitem{lecun1998gradient}
Y.~LeCun, L.~Bottou, Y.~Bengio, and P.~Haffner.
\newblock Gradient-based learning applied to document recognition.
\newblock {\em Proceedings of the IEEE}, 86(11):2278--2324, 1998.

\bibitem{jegou2011product}
H.~Jegou, M.~Douze, and C.~Schmid.
\newblock Product quantization for nearest neighbor search.
\newblock {\em IEEE Transactions on Pattern Analysis and Machine Intelligence},
  33(1):117--128, 2011.

\bibitem{gionis1999similarity}
A.~Gionis, P.~Indyk, and R.~Motwani.
\newblock Similarity search in high dimensions via hashing.
\newblock In {\em VLDB}, volume~99, pages 518--529, 1999.

\bibitem{charikar2002similarity}
M.~Charikar.
\newblock Similarity estimation techniques from rounding algorithms.
\newblock In {\em STOC}, pages 380--388. ACM, 2002.

\bibitem{ailon2009fast}
N.~Ailon and B.~Chazelle.
\newblock The fast {J}ohnson--{L}indenstrauss transform and approximate nearest
  neighbors.
\newblock {\em SIAM Journal on Computing}, 39(1):302--322, 2009.

\bibitem{gong2012iterative}
Y.~Gong, S.~Lazebnik, A.~Gordo, and F.~Perronnin.
\newblock Iterative quantization: A procrustean approach to learning binary
  codes for large-scale image retrieval.
\newblock {\em IEEE Transactions on Pattern Analysis and Machine Intelligence},
  35(12):2916--2929, 2012.

\bibitem{heo2015spherical}
J.~Heo, Y.~Lee, J.~He, S.~Chang, and S.~Yoon.
\newblock Spherical hashing: Binary code embedding with hyperspheres.
\newblock {\em IEEE Transactions on Pattern Analysis and Machine Intelligence},
  37(11):2304--2316, 2015.

\bibitem{kong2012isotropic}
W.~Kong and W.~Li.
\newblock Isotropic hashing.
\newblock In {\em NIPS}, pages 1646--1654, 2012.

\bibitem{omondi2006fpga}
A.~Omondi and J.~Rajapakse.
\newblock {\em FPGA Implementations of Neural Networks}, volume 365.
\newblock Springer, 2006.

\bibitem{pehlevan2018similarity}
C.~Pehlevan, A.~Sengupta, and D.~Chklovskii.
\newblock Why do similarity matching objectives lead to hebbian/anti-hebbian
  networks?
\newblock {\em Neural Computation}, 30(1):84--124, 2018.

\end{thebibliography}

\end{document}